\def\year{2021}\relax
\documentclass[letterpaper]{article} 
\usepackage{aaai21}  
\usepackage{times}  
\usepackage{helvet} 
\usepackage{courier}  
\usepackage[hyphens]{url}  
\usepackage{graphicx} 
\usepackage{natbib}  
\usepackage{caption} 
\usepackage{cite}
\usepackage{tabularx}
\usepackage{algorithm}
\usepackage{algpseudocode}
\usepackage{amsmath}
\usepackage{subcaption} 
\usepackage{verbatim }
\graphicspath{{.}{images/}}
\usepackage{diagbox} 
\usepackage{hyperref}

\makeatletter

\newcommand{\Rmnum}[1]{\expandafter\@slowromancap\romannumeral #1@}
\makeatother

\title{A Two-stage Multi-modal Affect Analysis Framework for Children with \\ Autism Spectrum Disorder}
\author{

    Jicheng Li, Anjana Bhat, Roghayeh Barmaki
}
\affiliations{

    University of Delaware\\

    \{lijichen, abhat, rlb\}@udel.edu

}

\begin{document}

\maketitle

\begin{abstract}
Autism spectrum disorder (ASD) is a developmental disorder that influences communication and social behavior of a person in a way that those in the spectrum have difficulty in perceiving other people's facial expressions, as well as presenting and communicating emotions and affect via their own faces and bodies. 
Some efforts have been made to predict and improve children with ASD's affect states in play therapy, a common method to improve children's social skills via play and games. However, many previous works only used pre-trained models on benchmark emotion datasets and failed to consider the distinction in emotion between typically developing children and children with autism.
In this paper, we present an open-source two-stage multi-modal approach  leveraging acoustic and visual cues to predict three main affect states of children with ASD's affect states (positive, negative, and neutral) in real-world play therapy scenarios, and achieved an overall accuracy of $72.40\%$. 
This work presents a novel way to combine human expertise and machine intelligence for ASD affect recognition by proposing a two-stage schema. 

\end{abstract}

\section{Introduction}
Autism is the fastest-growing developmental disorder in the United States: approximately 1 in 54 children is on the autism spectrum \cite{baio2018prevalence}. Individuals with ASD are characterized by having significant social communication impairments such as inefficient use of social gaze, gestures, and verbal communication \cite{NIH2018ASD}. Thus, individuals in the spectrum have difficulty perceiving and presenting communication cues such as emotion. Previous research has shown that play therapy can improve children's social and emotional skills and perceive their internal emotional world better \cite{chethik2003childrenTherapy}. The video recordings of play therapy interventions can provide a rich source to analyze children's emotion or affect states during treatment sessions.

\begin{figure}[h]
    \centering
    \includegraphics[width = \columnwidth]{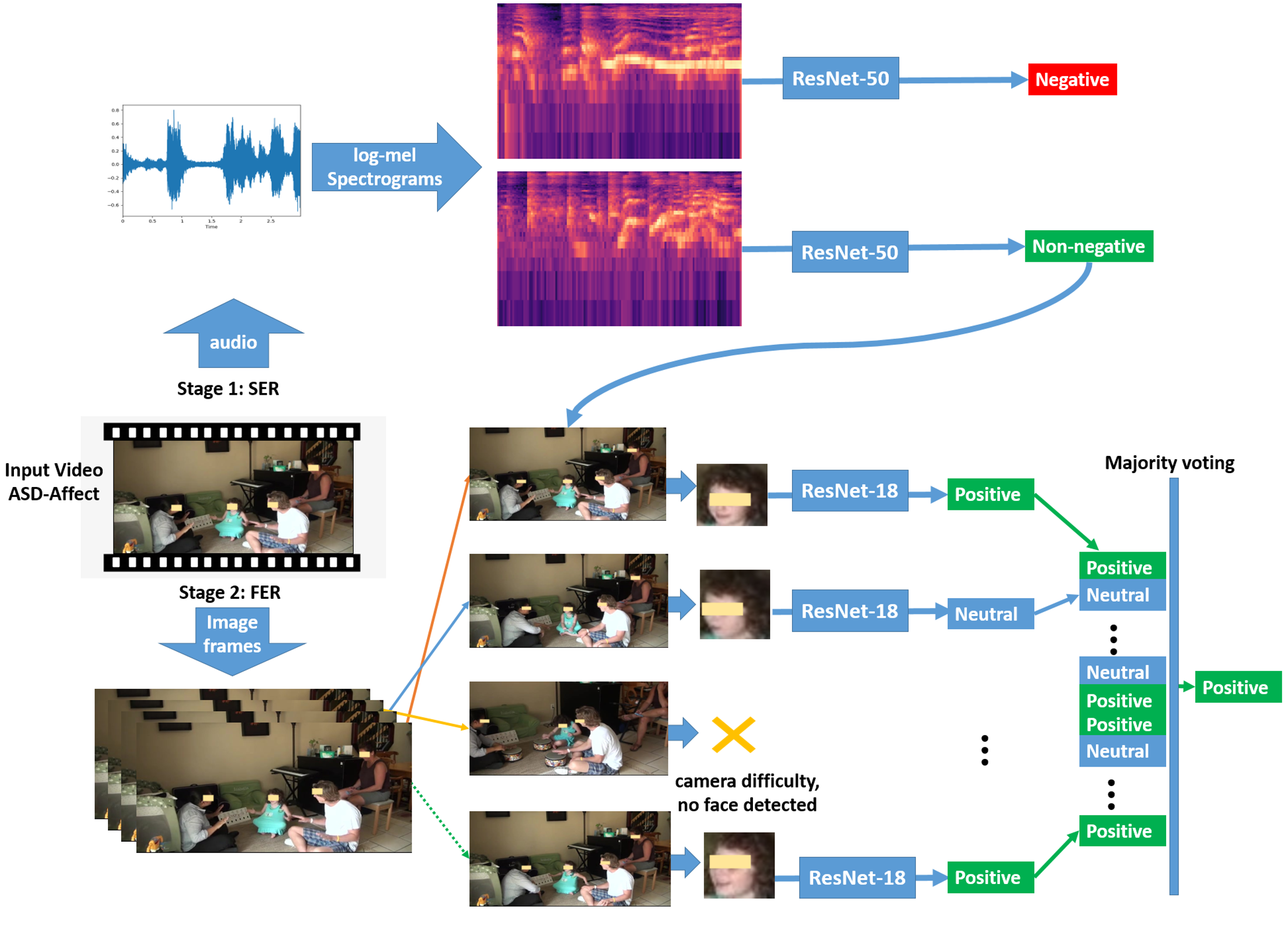}
    \caption{The workflow of our proposed affect prediction framework atop ASD-Affect dataset. Distinct emotional habits of ASD children inspire the two-stage schema, as they tend to scream and shout more in negative affect states and smile more in positive ones. Given a video input, we first classify negative data via speech emotion recognition (SER). Non-negatives will be passed to the second stage, where the model will decide whether it's positive or neutral based on facial expressions.}
    \label{fig:workflow}
\end{figure}
In this paper, we present a two-stage affect prediction method using video data (see Figure 1). We use a subset of ASD-affect dataset from \cite{Anjana2019Therapy} which includes more than four different therapeutic games for children to play. Sample settings of ASD-affect are shown in Figure 2. 

Emotion recognition is the process of identifying human emotion by multiple cues, including facial or spoken expressions, physiological and biological signals. Facilitated by machine learning techniques, computer vision, speech and signal processing, we can automate the process of emotion recognition. Researchers have shown that messages pertaining to feelings, affects and attitudes of interpersonal communication significantly reside in facial expressions and speech \cite{mehrabian1971silent, Dhall2012AFEW}. Inspired by that, in this paper, we define the problem as automating emotion recognition for children with ASD using multi-modal inputs, especially from visual and audio signals.
\newline
\newline

\begin{figure}[htb]
    \centering
    \includegraphics[width = \columnwidth]{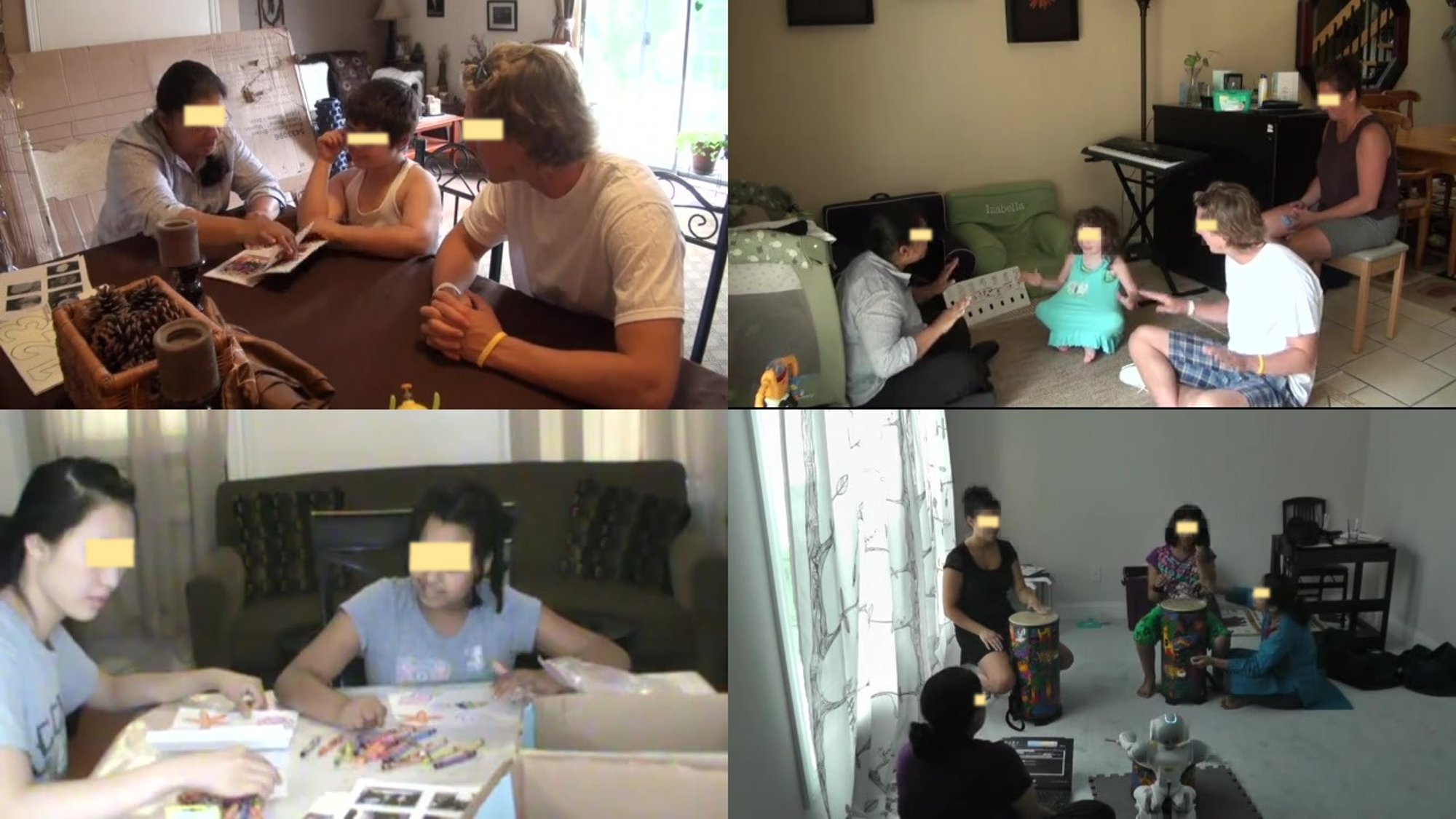}
    \caption{Sample data of ASD-affect. ASD-affect is featured by rich diversities in subjects and scenarios. Children with ASD were conducting various activities under therapists' instructions. These activities include singing, dancing, drumming, yoga, and interaction with robots. The dataset consists of more than 20 hours of therapy recordings.}
    \label{fig:asd_affect_scenes}
\end{figure}

However, there are several challenges in the process of automatic emotion recognition on ASD-affect dataset:

\textbf{Insufficient public dataset}: having adequate labeled training data that include as many variations of the populations and environments as possible is important for the design of a deep expression recognition system. However, due to privacy concerns, ASD dataset, especially from children is very scarce. ASD-related multi-modal dataset, which records children's behaviors in play therapy, is even more sparse.
    
    
 \textbf{Domain shifts}: existing methods \cite{Doyran2019PlayTherapy} have directly applied pre-trained model of normal people for play therapy analysis, either explicitly or implicitly under the assumption that emotional traits such as facial expressions of typically developing people and ASD children are the same or similar. We argue that such simplifications are not always appropriate given that ASD children suffer from affect and communication disorder and cannot express their emotions appropriately, especially for children diagnosed at level three \cite{Weitlauf2014levelsASD}.
    
\textbf{Data noise}: many existing benchmarks for emotion recognition \cite{Lucey2010CK+, valstar2010MMI, burkhardt2005EmoDB} are posed and collected in a controlled laboratory setting. In contrast, our dataset was collected in an in-the-wild manner featuring various backgrounds, people, activities, and durations. Therefore, ASD-affect has lots of noise, which requires substantial data cleaning and post-processing.  
    
\textbf{Sparse labeling}: unlike other multi-modal benchmark datasets \cite{Dhall2012AFEW, Nojavanasghari2016EmoReact} where the duration of data samples are in the scale of seconds (usually less than 5 seconds), examples of ASD-affect dataset may last for minutes, equivalent to lacking ground truth or introducing excessive noise to the dataset. This is because benchmark datasets were intentionally collected and labeled for autonomous recognition by machine, but ASD-affect was initially compiled and annotated to serve human experts.
    
Despite all these challenges, we used transfer learning, fine-tuning, and data post-processing - listed in the following sections - to prepare ASD-affect for further analysis using speech and facial emotion recognition methods.  

In this paper, we propose a two-stage framework to evaluate  affect states of children in play therapy scenarios using multi-modal emotion clues. This method effectively combines prior knowledge from human experts with machine intelligence. To distinguish children between three different affect states - neutral, positive, and negative - in stage 1, the model predicts whether children are in a negative state based on negative symptoms (shouting and screaming) residing in speech. In stage 2, children's emotions in positive and neutral states are recognized by distinct facial expressions. The workflow of our framework is presented in Figure \ref{fig:workflow}. Our approach enables physical therapists to better and more efficiently analyze the effectiveness of play therapy interventions since human professionals require a fair amount of training to better understand the behaviors and emotional states of ASD children. This method can be further applied to data annotation and label verification for other ASD datasets, as actions of ASD children resemble relatively well.

This paper is organized as follows. We first summarize related works in emotion recognition and play therapy analysis in Section 2. Section 3 describes the method we proposed, followed by experiment and results discussion presented in Section 4, 5 and 6. Lastly, section 7 outlines the conclusion and future steps for this research.

\section{Related Work}
\subsection{Multi-modal Emotion Recognition}
Emotions are convoluted psychological states composed of several components: personal experience, physiological, behavioral, and communicative reactions. There are two mainstream emotion representations: discrete model \cite{ekman1994strong} and dimensional model. In this paper, we used discrete emotion models.

Emotions can be carried in various modalities of inputs. Mehrabian shows that $55 \%$ of messages pertaining to feelings and attitudes of interpersonal communication is in facial expressions \cite{mehrabian1971silent}. Besides, Dhall suggests that audio modality can bring extra gain in emotion recognition accuracy \cite{Dhall2012AFEW}. Thus, multi-modal emotion recognition approaches usually outperform unimodal ones. Two main sub-sets of multi-modal emotion recognition models are facial expression recognition (FER) and speech emotion recognition (SER), which are also the main focus of this work.

\subsubsection{Facial Expression Recognition} FER systems can be divided into two main categories based on the feature representations: static and dynamic. In static-based methods, the feature representation is encoded with only spatial information from a single image frame. In contrast, dynamic-based approaches consider temporal relations among contiguous frames in the input facial expression sequence \cite{li2018deepFER_survey}. Li  proposed a bi-modality method \cite{Li2019BiModality}, where convolutional networks (CNNs) were used to recognize static facial expressions while a bi-direction long short term Memory (Bi-LSTM) was employed to learn dynamic facial expression sequences extracted by CNNs. Liu also embodied facial landmarks in the FER system \cite{Liu2018Multi}. However, these works were conducted on benchmark datasets \cite{Dhall2012AFEW} where sequential relation of images is well-preserved so that sequential methods are able to function. Conversely, our ASD dataset was recorded in the natural or in-the-wild settings; so we could only use a static-based method to classify facial expressions in each frame, without considering temporal information.

\subsubsection{Speech Emotion Recognition}
Speech is a rich, dense form of communication that can convey information effectively. There are two classical ways to extract emotional features from speech. First is to obtain low-level discriminator features of speech, such as Mel-frequency cepstral coefficients \cite{InteractionAware2019Yeh,yoon2020attentive}. Another way is to convert audio to spectrograms then use CNNs as feature extractors \cite{Zhang2018Pyramid, zhao2019speech1D_2D_CNN_LSTM}. In this paper, we use spectrograms as audio representations.

\subsection{Play Therapy Analysis}
Play therapy is an approach to psychotherapy where a child is engaging in play activities. Doyran and colleagues \cite{Doyran2019PlayTherapy} proposed a visual and text-based framework to track the affective state of a child during a play therapy session. However, audio modality was less explored in their work, and categorical representations of facial expressions needed more investigation. Bangerter investigated the spontaneous production of facial expressions of individuals with ASD as a response to entertaining videos \cite{bangerter2020automated}. It turned out that individuals with ASD showed less evidence of facial action units relating to positive facial expression than typically developing children. Due to small face sizes and low resolution of ASD-affect dataset, 
using facial action units approach in the current work was not feasible, but we are looking into it in future.

\section{Method}
In this paper, we propose an open-source two-stage multi-modal framework to predict children's affect states in play therapy leveraging visual and audio information.
First, we distinguished negative videos from non-negative ones (neutral and positive) using spectrograms generated from audio. Next, to differentiate between positive and neutral videos, we used static-based facial expression recognition methods. The workflow of this method is illustrated in Figure \ref{fig:workflow}.

\subsection{Two-stage Schema}
Our data assessment on ASD-affect inspires the two-stage approach. Children who negatively and passively participated in play interventions tend to shout and scream more often, and such characteristic is manifest in speech. However, there are no significant diversities in speech emotion between neutral and positive recordings. Instead, children are smiling when positively engaged in therapy, while their facial expressions remain neutral more often in neutral states. Therefore, we chose to leverage the variance in facial expressions to distinguish between positive and neutral data.

\subsection{Stage 1: Negative vs Non-negative}
Since distinct speech emotions exhibit different patterns in the energy spectrum, to capture emotion features from speech, we selected log-Mel spectrograms which have been  effective in speech emotion recognition tasks in the past \cite{zhao2019speech1D_2D_CNN_LSTM, Zhang2018Pyramid, Chen20183D}. A spectrogram is a visual representation of the spectrum of a signal's frequencies as it varies with time. It is a graph with two geometric dimensions: time and frequency. The amplitude of a particular frequency at a particular time is represented by the intensity or color of each pixel in the spectrogram. A Mel-spectrogram is a spectrogram where the frequencies are converted to the Mel scale - a perceptual scale of pitches assessed by listeners to be equal in distance from one another \cite{stevens1937melscale}. We used the logarithmic form of Mel-spectrogram to better reflect emotions, since humans perceive sound in a logarithmic scale \cite{venkataramanan2019emotion}. 

\subsection{Stage 2: Neutral vs Positive}
As noted earlier, due to image resolution constraints,  temporal information was not well preserved as adjacent frames were discarded frequently in the data cleaning stage, causing sequential models to fail to converge. Therefore, we needed to use static-based methods that solely depend on one frame to predict facial expression. We choose ResNet-18 \cite{he2016ResNet} with a decreased input size to better fit the average face sizes detected in ASD-affect. We pre-trained the model using EmoReact \cite{Nojavanasghari2016EmoReact}, a multi-modal emotion dataset of children and fine-tuned it on ASD-affect dataset.

\section{Experiment}

\subsection{Data Processing}
\subsubsection{ASD-affect Dataset}
Bhat and colleagues proposed that use of embodied, multisystem interventions can enhance various social communication, perceptuo-motor, and cognitive-behavioral impairments of children with ASD \cite{Anjana2019Therapy}. They have studied the effects of various embodied creative interventions, including the themes of robotic, musical, physical activity, yoga, and dance therapy interventions for children with ASD. The video recordings of such interventions, known as the ASD-affect dataset, have provided a rich source for analyzing children's affect states in play therapy. In this paper, we used a subset of ASD-affect from six children. Sample data of ASD-affect are shown in Figure \ref{fig:asd_affect_scenes}.

\subsubsection{Data Reconstruction}
\begin{figure}[ht]
    \centering
    \includegraphics[width = 0.8\columnwidth]{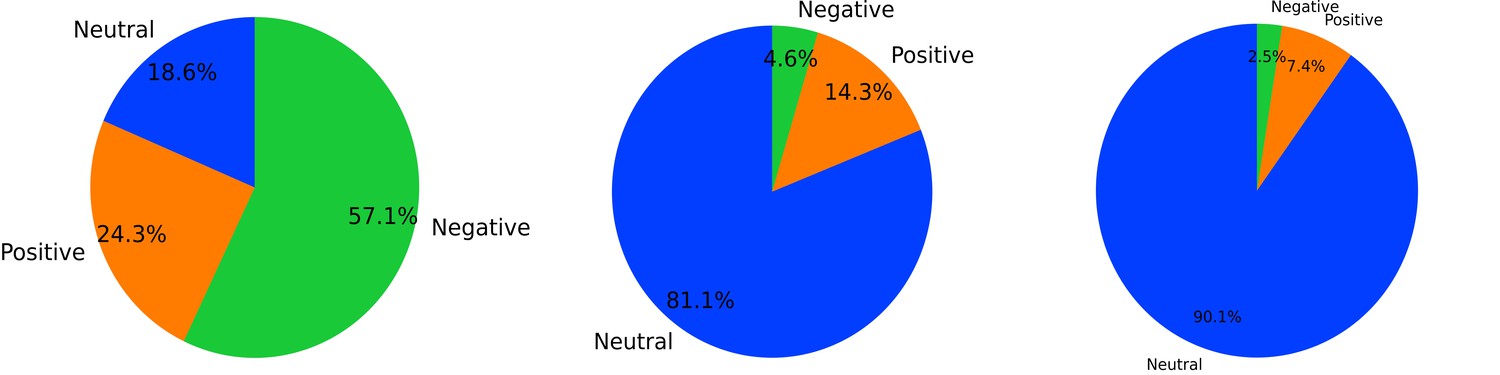}
    \caption{A comparison of ASD-affect and benchmark dataset AFEW \cite{Dhall2012AFEW}. Left: sample distribution of reconstructed AFEW. We relabeled fear, disgust, angry and sad as negative. Middle: data count distribution of ASD-affect. Right: data duration percentage of ASD-affect.}
    \label{fig:ASD_vs_AFEW}
\end{figure}

Originally, there were eight different types of labels in the ASD-affect: neutral, interested, positive, positive and talking, odd positive, runs away, camera difficulties, and negative. For our work, we reconstructed the dataset, and excluded - runs away and camera difficulties and odd positive labels - or merged some labels - interested, positive, positive and talking were all considered as positive labels. After this reconstruction step, we had a total of 471 clips from six children in three classes of positive (68 clips), neutral (384 clips), and negative (19 clips). Clips lengths were varied. See Figure \ref{fig:ASD_vs_AFEW} for reconstructed data distribution. 
 
\subsubsection{Log-Mel Spectrograms}
We first extracted audio tracks from video recordings. Audio files were stored in Waveform Audio File Format to retain high fidelity. We then applied noise reduction to audio files and removed silent utterances. Afterwards, we split each audio file into equal-length segments of 3 seconds, and zero-padding was applied to the utterances whose duration is less than 3 seconds. We set this sequence length since the average audio duration in selected benchmark datasets for SER was 3 seconds \cite{burkhardt2005EmoDB, livingstone2018RAVDESS}. After that, log-Mel spectrograms were generated from each audio segment using \textit{librosa} toolkit \cite{mcfee2015librosa}.  We set the Fast Fourier Transform (FFT) window length and hop length to 2,048 and 512, respectively. 64 Mel bands were used in the spectrogram generation. A total of 9,968 log-Mel spectrograms were generated, including 134 negative samples and 9,834 non-negative samples. We then used down-sampling on non-negative spectrograms to even out data, and reduce data imbalance. 
In both the training and testing phase, all log-Mels were normalized by the global mean and standard deviation of the training set. All spectrograms were resized to $224 \times 224$ to match the network's input size.   
\subsubsection{Facial Images}
We first extracted image frames from raw video clips at a specific sampling rate. Considering that duration of neutral clips were typically longer than positive ones in ASD-affect, we set the sampling rate to 3 frames per second (FPS) for positive video clips and 1 FPS for the neutral to stratify data proportionally. Then we used MTCNN \cite{Zhang2016MTCNN} to detect human faces in each frame. We selected 1,756 template faces of children (about 2\% of the total detected faces) to create a facial expression database for ASD-affect, consisting of 1,159  neutral and 706 positive faces. Each selected face was manually labeled as either neutral or positive based on facial expressions. The children face dataset served as training, validation, and test set for the FER model used in the second stage via 5-fold cross-validation. We used random crop, rotation, shifting, illumination adjustment, and normalization techniques for data augmentation and noise reduction. Before training, all facial images were resized to $48 \times 48$ offline, then random cropped to $44 \times 44$ on-the-fly during training. In testing, faces were directly resized to $44 \times 44$.

Detected faces may belong to children, other persons in the scene or due to noise. To localize children's faces properly, we leveraged the children's face dataset to create a face embedding database, where each face was encoded as a 128-dimensional vector. Whenever a new face is encountered, we can compare its embedding with the embedding database we have established to find matches. A 'match' is defined as the cosine similarity between the new face embedding and a known face embedding is less than a given confidence threshold. Only matched faces were used for predictions, and unmatched faces were excluded.

\subsection{Speech Emotion Recognition}
Since the whole dataset is imbalanced, where negative video clips are much less than non-negative ones, we applied weighted sampling to enhance negative samples' occurrence while working with spectrograms. We chose a batch size of 32, and the network was trained for 25 epochs. We used Adam \cite{kingma2014adam} optimizer, and the learning rate was set to 0.001.


\subsection{Facial Expression Recognition} 
The training set was the selected ASD children's faces, as mentioned above. We set the batch size to 64 for training while the total training epochs was 25, and chose Adam \cite{kingma2014adam} optimizer with an initial learning rate of 0.001. The learning rate was decreased by a factor of 0.1 every 20 epochs. Unlike the training phase, in testing, input images were captured every five frames from videos on-the-fly. Note that inputs in testing were not face crops but image frames, indicating that a face detector has to be applied to capture human faces from frames. MTCNN was then applied to test images to capture human faces. Detected faces were compared with the established children's face database. Once children's faces were matched and located, a trained model predicted children's facial expressions, and such predictions were considered valid votes. Frames were discarded if no target children's faces were detected, including no faces or only faces from others (e.g., therapists or parents). At the end of each video, for all the valid votes, if the portion of positive predictions exceeds a certain threshold, the whole video is predicted as positive, otherwise neutral. In this experiment, we set the threshold to 0.5, equivalent to majority voting. The workflow of the test phase is explained in Algorithm \ref{alg:FER}.
\newline
\newline

\begin{algorithm}[htb]
\caption{\textbf{Stage\_2\_testing(input video $v$, face detector $face\_det$ , children face embedding $child\_embeds$, classifier $model$, threshold T)}}
\label{alg:FER}
\begin{algorithmic}[1]
    \State $vote\_list = []$ \Comment{Initialize a list to store valid votes}
    \State Load video $v$, capture one image as input every 5 frames. All captured images are stored in $frame\_list$
    \For {$img$ in $frame\_list$}
        \State Detected face list $fl$ =  $face\_det (img)$
        \For{each face $f$ in $fl$}
            \If{$f \in child\_embeds$} \Comment{Locate children}
                \State $pred$ = $model(f)$
                \State $vote\_list.append(pred)$ 
            \EndIf
        \EndFor
    \EndFor
    
    \State $pos\_ratio = \frac{count(p_i == Positive) \text{\ in \ } vote\_list }{ vote\_list.length}$
    \If{$pos\_ratio>= T$} \Comment{Majority voting}
        \State return Positive
    \Else
        \State return Neutral
    \EndIf
\end{algorithmic}
\end{algorithm}


\section{Results}
We used 5-fold cross-validation to report findings from our participant videos (recordings from two children were merged together due to small number of video clips, totalling five batches of data from six children. Since we had imbalanced classes, in addition to accuracy, we reported recall, F1 score, G-mean value, and ROC-AUC score for more in-depth analysis. 

\subsection{Stage 1: Negative vs. Non-Negative}
\begin{figure}[htb]
    \centering
    \includegraphics[width= \columnwidth]{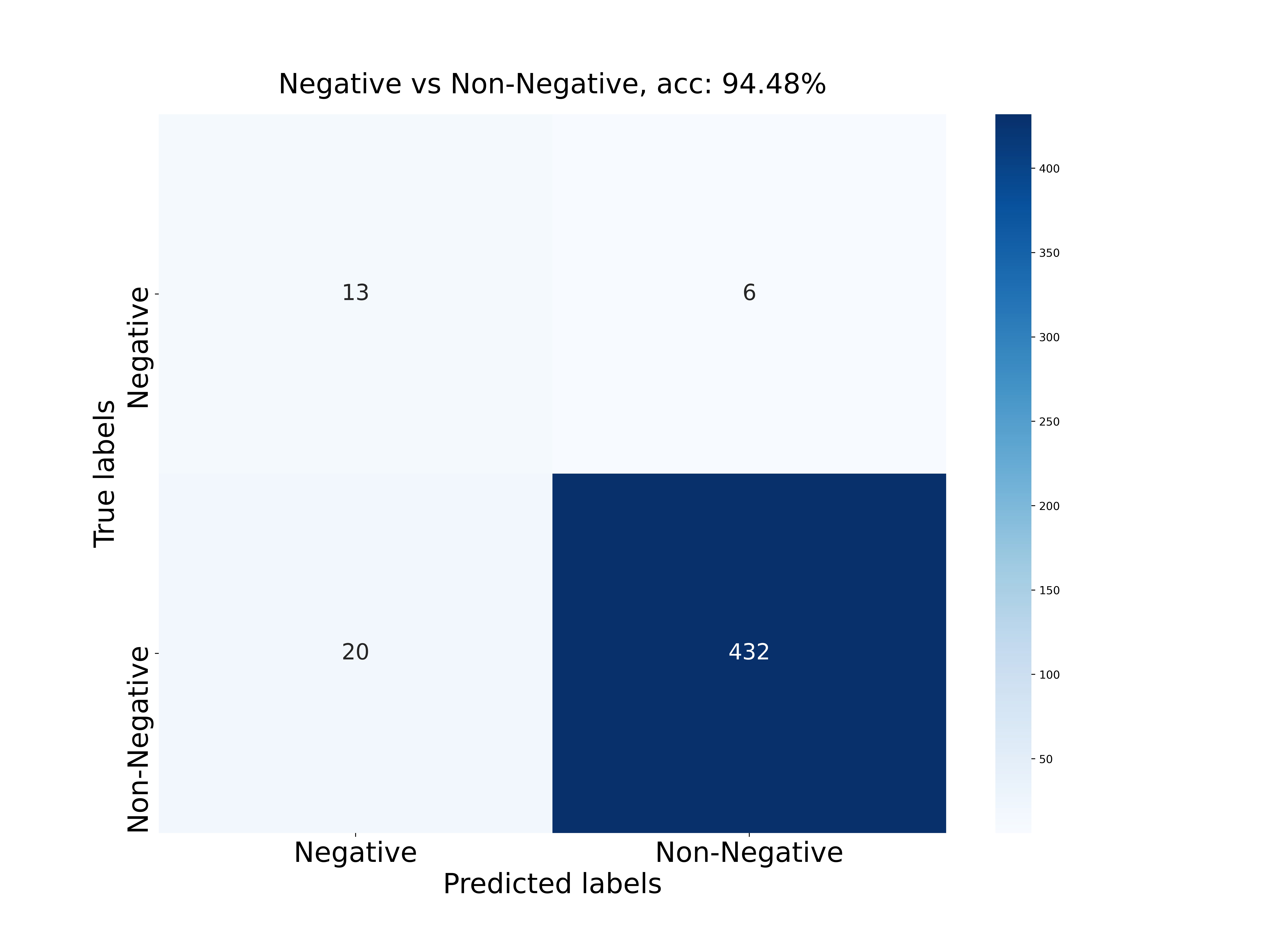}
    \caption{Confusion matrix for prediction of negative and non-negative videos extracted from audio modality and SER.}
    \label{fig:cm_neg}
\end{figure}

\begin{figure}[htb]
    \centering
    \includegraphics[width= \columnwidth]{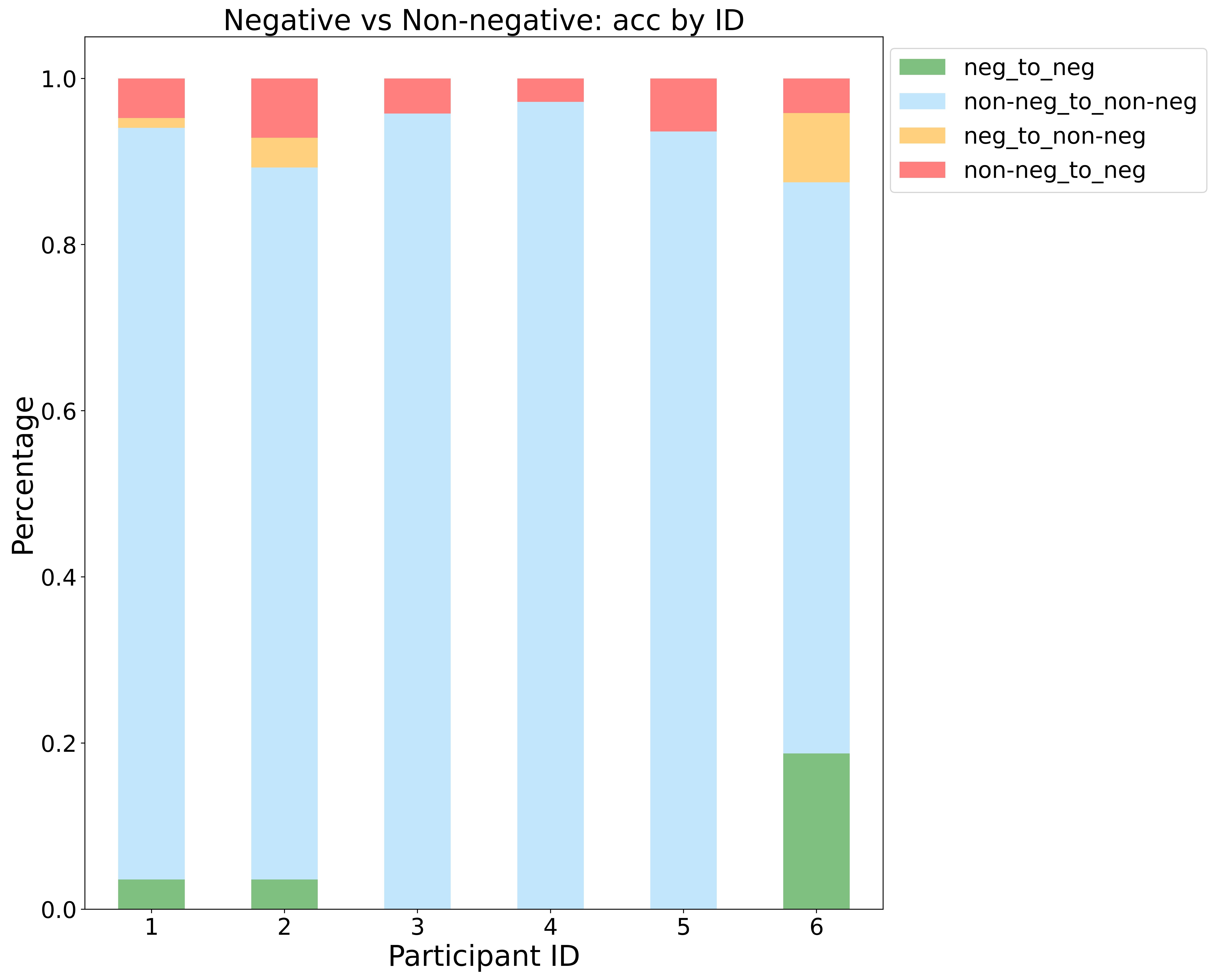}
    \caption{Classification results per participant in stage 1, negative  vs.  non-negative. Y-axis  shows  the  percentages  of  predicted labels for video clips from every participant (data is normalized based on total video clips from every participant, because video counts vary significantly). The legend format is: true label to predicted label (e.g. $neg\_to\_non-neg$ means a negative video is misclassified as a non-negative one).}
    \label{fig:id_acc_SER}
\end{figure}

We achieved an accuracy of $94.48\%$ and an F1 score of $0.97$. The recall of negative and non-negative labels is $68.42 \%$ and $ 95.57\%$. Besides, the G-mean value and ROC-AUC score are 0.92 and 0.93, respectively. The confusion matrix of stage 1 is shown in Figure \ref{fig:cm_neg}. The classification results from recordings of each participant is shown in Figure \ref{fig:id_acc_SER}. 



\subsection{Stage 2: Neutral vs. Positive}
\begin{figure}[htb]
    \centering
    \includegraphics[width= \columnwidth]{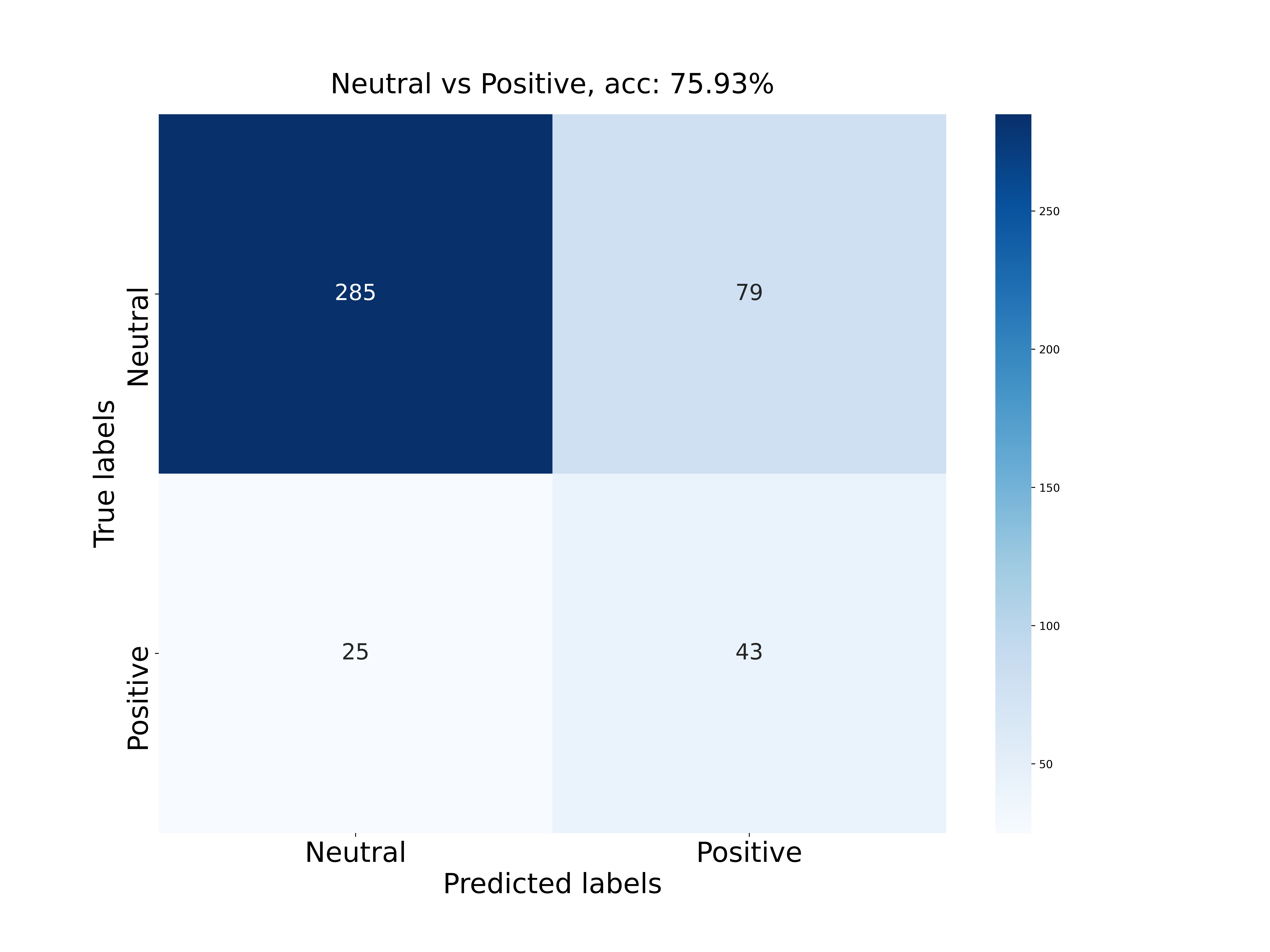}
    \caption{Confusion matrix for prediction of positive and neutral videos extracted from FER.}
    \label{fig:cm_pos_neu}
\end{figure}

\begin{figure}[htb]
    \centering
    \includegraphics[width= \columnwidth]{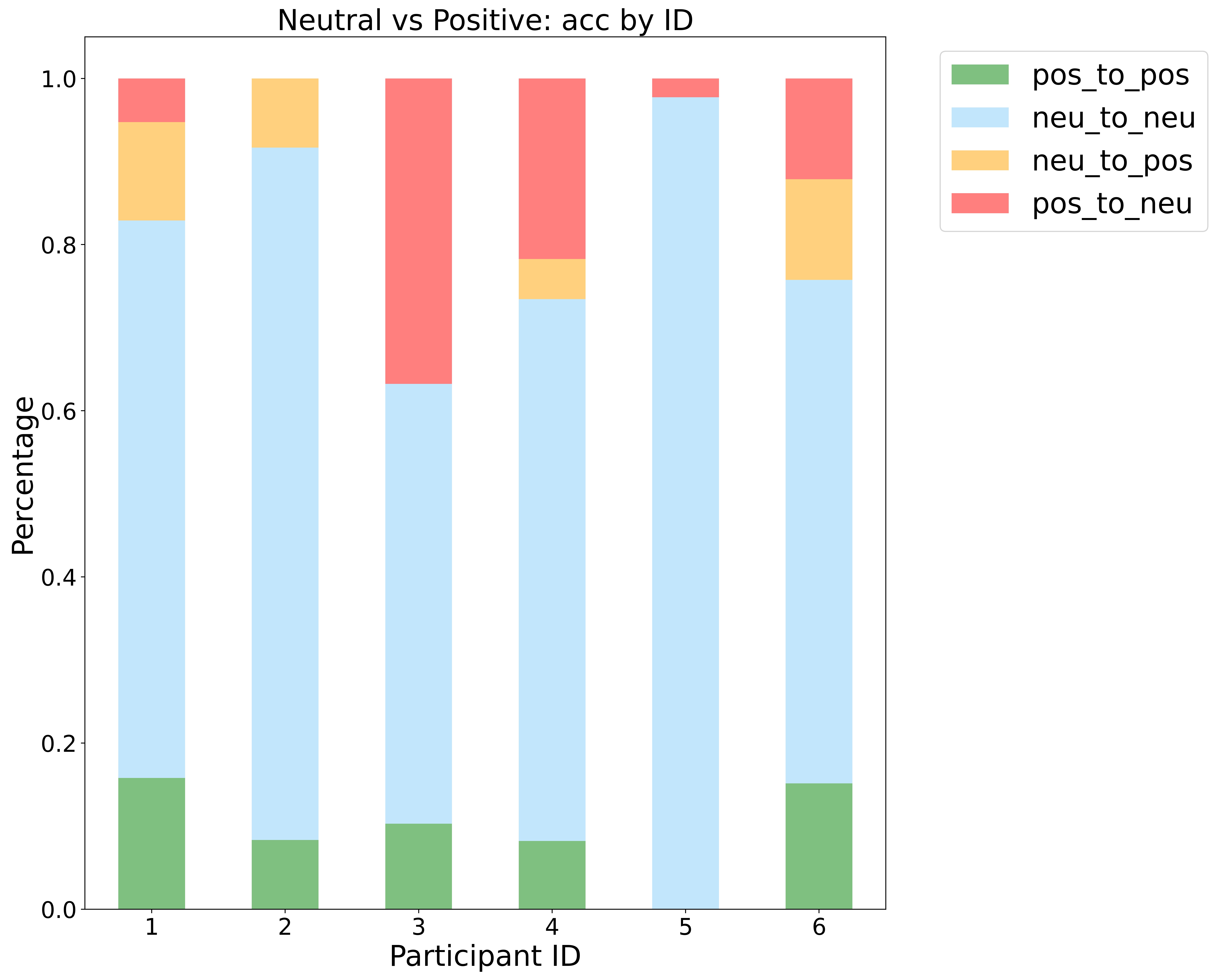}
    \caption{Classification results per participant in stage 2, neutral vs. positive. Y-axis shows the percentages of predicted labels for video clips from every participant (data is normalized based on total video clips from every participant, since video counts vary significantly). The legend format is: true label to predicted label (e.g. $neu\_to\_pos$ means a neutral video is misclassified as a positive one). Blue and green represent correct predictions for positive and neutral video clips, while red and orange stand for miss-classified ones. The more considerable sum of green and blue areas are, the better accuracy our model achieves. This figure illustrates that our model generalizes well among all individuals in ASD-affect in distinguishing neutral and positive videos.}
    \label{fig:id_acc_FER}
\end{figure}

We reached an overall accuracy of $75.93\%$, where recall for neutral and positive class was $78.29 \%$ and $63.24 \%$, respectively. Confusion matrix is shown in Figure \ref{fig:cm_pos_neu}. Also, F-1 score is calculated as 0.79. The results from each participant's videos are shown in Figure \ref{fig:id_acc_FER}. 


\subsection{Overall Accuracy}
\begin{figure}
    \centering
    \includegraphics[width= \columnwidth]{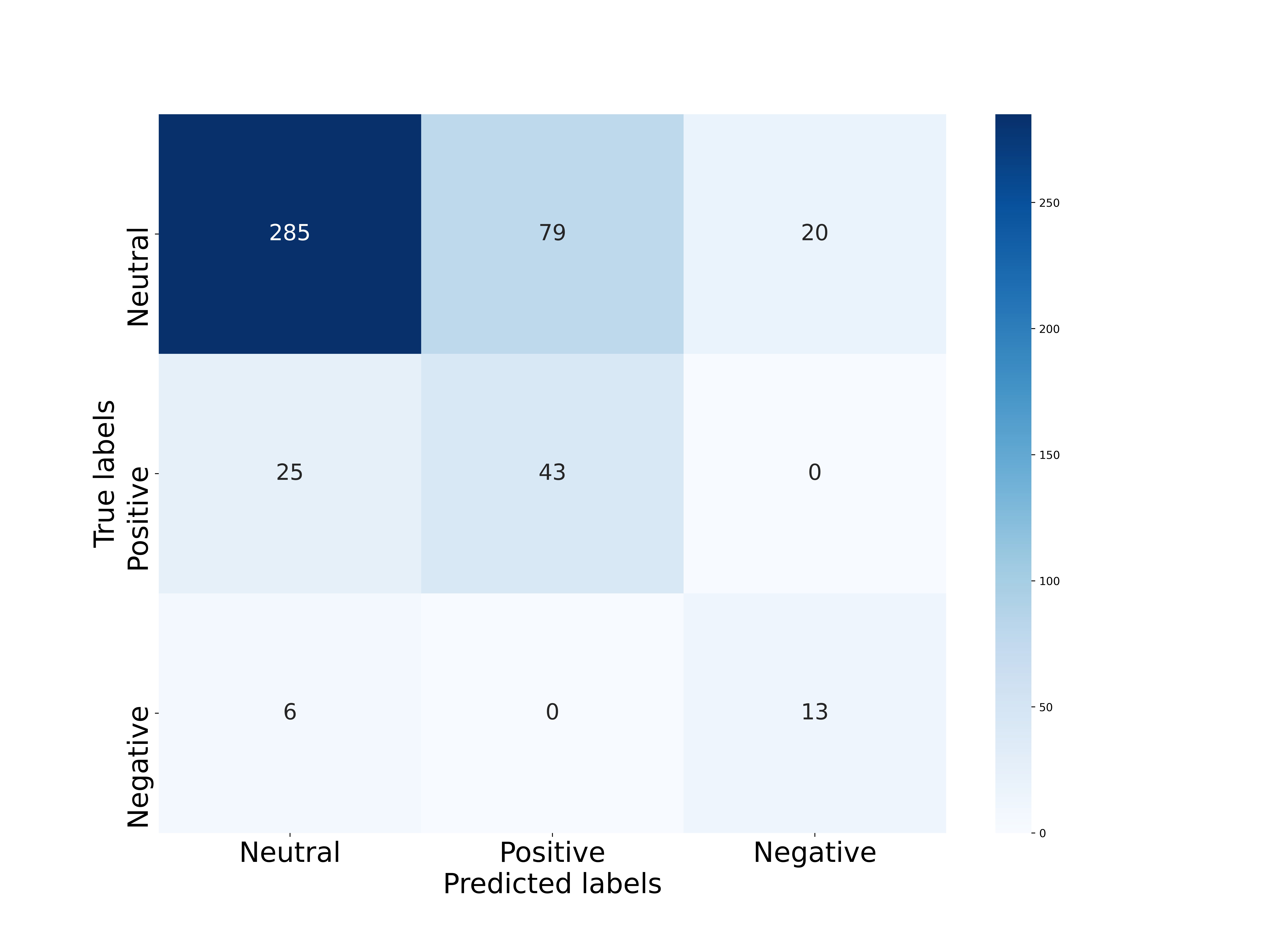}
    \caption{Confusion matrix for three classes.}
    \label{fig:cm_overall}
\end{figure}

According to the confusion matrix for all three classes (Figure \ref{fig:cm_overall}), we have correctly classified 285 neutral videos, 43 positive videos, and 13 negative ones. Adding them up, we had 341 out of 471 precisely classified clips, leading to an overall accuracy of  $72.40\%$ and an F1 score of $0.75$.
\newline
\newline

\section{Discussion}
Overall, our method achieves an acceptable performance in both stages. However, there is a noticeable accuracy gap between stage 1 and stage 2 and between dominant and non-dominant classes of each stage. If we consider our problem a two binary classification problems --stage one with negative vs non-negative samples, and stage two of non-negative samples classified to positive and neutral samples-- for non-dominant labels in both stages, which are negatives in stage 1, and positives in stage 2, the recall rates are very comparable: $68.42\%$ and $63.24 \%$ respectively.
On the other hand, for both of dominant classes in stages 1 and 2, the recall for non-negatives is significantly higher than neutral labels ($95.57\%$ vs. $78.29\%$). This may because speech emotion features, such as shouting and screaming, are more distinct and recognizable and describe negative videos better. Moreover, the distinction of positive from neutral labels in stage 2 was very tough even for subject matters experts due to data noise and low video resolution. As such, SER performed relatively better than FER for our ASD-Affect dataset.

\section{Conclusion}
This paper proposed a novel framework for automatic emotion recognition of children with ASD using multi-modal information (facial and speech emotion), providing a baseline model to affect states analysis in play therapy. This work also has implications on automated affect annotation for play therapy video recordings. Besides, the framework leverages human expertise to a great extent by proposing a two-stage schema, a novel way to combine human knowledge and machine intelligence in ASD-related research. 

We anticipate expanding this project in the future in multiple directions. We aim to offer a semi-automated annotation framework to assist subject-matter experts swiftly annotating the recordings from children with autism. We discussed some challenges of the pre-recorded videos in our dataset, especially the low-resolution issue. To overcome the problem, we plan to collect more audio-visual data with higher resolution to deploy other FER techniques, including sequential and action-unit based approaches mentioned in the paper. Furthermore, deficits in mutual gaze, and shared gaze is also known as a strong predictor of autism among children \cite{zhao2017atypical}, which we are interested in investigating in the future, as a next line of our previous research \cite{Guo_Barmaki_2020} on automatic detection of mutual gaze among adults. 

\section{Acknowledgments}
We wish to acknowledge the support from the entire research team, study participants and their caregivers to collect ASD-affect dataset. We also thank our sponsor, Amazon Research Awards Program for the generous support. Any opinions, findings, and conclusions or recommendations expressed in this material are those of the authors and do not necessarily reflect the views of the sponsors.

\bibliography{ref}{}
\end{document}